\documentclass{article}
\usepackage{ijcai22}

\usepackage{times}

\usepackage{soul}
\usepackage{url}
\usepackage[hidelinks]{hyperref}
\usepackage[utf8]{inputenc}
\usepackage[small]{caption}
\usepackage{graphicx}
\usepackage{amsmath}
\usepackage{booktabs}
\usepackage{times}
\usepackage{epsfig}
\usepackage{bbding}
\usepackage{pifont}
\usepackage{wasysym}
\usepackage{amssymb}
\usepackage{subfigure}
\usepackage{multirow}
\usepackage{algorithmicx,algorithm}
\usepackage{mathrsfs}
\usepackage{verbatim}
\usepackage{comment}
\usepackage{color}
\usepackage{textcomp}
\usepackage{stfloats}
\usepackage{newfloat}
\usepackage{listings}
\usepackage{threeparttable}
\title{ACTIVE:Augmentation-Free Graph Contrastive Learning for Partial Multi-View Clustering}

\author{
Yiming Wang$^{1,2}$\footnote{Contact Author}\and
Dongxia Chang$^{1,2}$\and
Zhiqiang Fu$^{1,2}$\And
Jie Wen$^{3,4}$\And
Yao Zhao$^{1,2}$\\
\emails
\{wangym,dxchang,zhiqiangfu,yzhao\}@bjtu.edu.cn,
jiewen\_pr@126.com
}

\begin{document}

\maketitle
\begin{abstract}

In this paper, we propose an augmentation-free graph contrastive learning framework, namely ACTIVE, to solve the problem of partial multi-view clustering. Notably, we suppose that the representations of similar samples (i.e., belonging to the same cluster) and their multiply views features should be similar. This is distinct from the general unsupervised contrastive learning that assumes an image and its augmentations share a similar representation. Specifically, relation graphs are constructed using the nearest neighbours to identify existing similar samples, then the constructed inter-instance relation graphs are transferred to the missing views to build graphs on the corresponding missing data. Subsequently, two main components, within-view graph contrastive learning (WGC) and cross-view graph consistency learning (CGC), are devised to maximize the mutual information of different views within a cluster. The proposed approach elevates instance-level contrastive learning and missing data inference to the cluster-level, effectively mitigating the impact of individual missing data on clustering. Experiments on several challenging datasets demonstrate the superiority of our proposed methods.
\end{abstract}

\section{Introduction}

Multi-view Clustering (MVC), as one of the most important unsupervised multi-view learning tasks, has attracted increasing attention during the last few decades. The pursuit of MVC is to improve clustering performance by exploiting complementary and consensus information from multiple views. Towards this goal, a variety of related methods have been proposed \cite{conf/icml/0001HLZZ19,journals/pami/ZhangFHCXTX20,conf/aaai/YinHG20}. A common assumption for most MVC methods is that all views are complete. However, in some practical applications \cite{journals/isci/ChaoSLWLLB19}, the absence of partial views makes most multi-view clustering methods inevitably degenerate, which raises a challenging case as partial multi-view clustering (PMVC). 

Many efforts have been devoted to addressing PMVC in recent years, which can be roughly divided into four categories: Non-negative Matrix Factorization (NMF) based methods, kernel-based methods, spectral clustering based methods, and Deep Neural Networks (DNNs) based methods. In brief, NMF based methods \cite{conf/aaai/LiJZ14,conf/ijcai/ZhaoLF16} learn a common latent space for complete samples and private latent representations for incomplete samples. Li et al. \cite{conf/aaai/LiJZ14} first proposed NMF based partial multi-View clustering method to learn a latent subspace over two views. Motivated by this method, a lot of partial multi-view clustering approaches \cite{conf/ijcai/HuC18,conf/aaai/WenZ0ZFL19} extend NMF to data with more views. Kernel based methods \cite{conf/ijcai/WangLZTLHXY19,journals/pami/LiuZLWTYSWG19} leverage kernel matrices of the complete views for completing the kernel matrix of incomplete view. To reduce the storage and computational complexity of kernel learning, Liu {\em et al.} \cite{journals/pami/LiuLTXXLKZ21} propose a more efficient approach with linear computational complexity. As a typical spectral clustering based method, perturbation-oriented incomplete multi-view clustering (PIC) \cite{conf/ijcai/WangZLYZ19} first learns the consensus Laplacian matrix using a consistent Laplacian graph based on perturbation theory. These three types of methods show satisfactory performance but suffer from two main disadvantages: (1) These models are generally shallow models with limited capacity to reveal the relations in complex multi-view data; (2) It is inefficient to employ them for large-scale and high-dimensional datasets. 

Many DNN-based PMVC methods have emerged recently thanks to the amazing power of DNNs to handle multi-view data. These DNN-based methods learn common representations for clustering via Autoencoders (AEs) with techniques to mitigate the effects of absences. Among them, some methods \cite{conf/ijcai/XuGZWNL19,9258396,journals/tip/WangDTGF21} utilize generative adversarial networks (GAN) \cite{journals/corr/GoodfellowPMXWOCB14} to guide the imputation of missing data based on the complete views. Additionally, to reduce the impact of missing views, the graph embedding strategy is utilized in some methods \cite{conf/ijcai/WenZ0ZFX20,conf/mm/0001ZZWFXZ20}. Besides, \cite{conf/cvpr/0001GLL0021} proposes a joint representation learning and data recovery framework based on contrastive learning. 

These DNN-based methods achieve reasonable performance in simultaneously inferring missing data and learning representations. However, it is more valuable for clustering tasks to learn common characteristics within clusters than individual characteristics, which can lead to a more aggregated within-cluster and more dispersed between-cluster latent space representation. Meanwhile, the missing inferences and representation learning of individuals are more sensitive to missingness. Thus it is appropriate to perform similar inference and learn similar representations for samples from the same cluster. 

On this basis, we design an \textbf{A}ugmentation-free graph \textbf{C}ontras\textbf{TIVE} learning framework (ACTIVE) to solve the challenge of PMVC. Unlike existing deep partial multi-view clustering approaches, we focus on lifting the instance-level inter-view consistency to the cluster-level inter-view consistency. We assume that the representation of samples of one cluster from all views should be similar, which can effectively reduce the impact of missing views. To identify similar samples, relation graphs are constructed using the nearest neighbours on the existing samples. Since different features of the same instance should follow a high-dimensional semantic consistency \cite{conf/cvpr/0001GLL0021}, known inter-instance relation graphs can be transferred to the missing views to build graphs on the missing data. The main contributions of our method are summarized as follows:

\begin{itemize}
    \setlength{\itemsep}{1pt}
    \setlength{\parsep}{1pt}
    \setlength{\parskip}{1pt}
    \item We propose a new partial multi-view clustering method, ACTIVE, which simultaneously infers missing data and learns representations via cluster-level inter-view consistency. The cluster-level operations can effectively mitigate the impact of missing views.
    \item Different from other PMVC methods, in ACTIVE, relation graphs are transferred to the missing views to infer the missing samples, which can efficiently reduce the influence of missing views. 
    \item Two main modules, i.e., within-view graph contrastive learning and cross-view graph consistency learning, are devised to explore intra-cluster consistency and inter-view consistency.
    \item Our extensive experiments on some benchmark datasets clearly show that ACTIVE outperforms the state-of-the-art PMVC methods. We also conduct a series of ablation studies to validate the effectiveness of each objective in ACTIVE.
\end{itemize}

\section{Notations and Preliminaries}

\paragraph{Notations.} Formally, the partial multi-view data is denoted by $\mathcal{X} = \{X^{(1)},...,X^{(V)}\}$, where $X^{(v)} = \{x^{(v)}_1,...,x^{(v)}_N\} \in \mathbb{R}^{ N \times {d_v} }$, $V$ is the number of views, $N$ is the number of samples, $d_v$ is the feature dimension of $v$-th view, and elements of the missing samples are denoted as ‘NaN’. For convenience, matrix $U \in \mathbb{R}^{ N\times V}$ is used to record the available and missing condition of these $N$ samples, where $U_{i,v} = 1$ means the $i$-th sample is available in the $v$-th view, otherwise $U_{i,v} = 0$. Our goal is to group all the samples into $C$ clustering.

\begin{figure}[t]
	\centering
	\includegraphics[width=7cm]{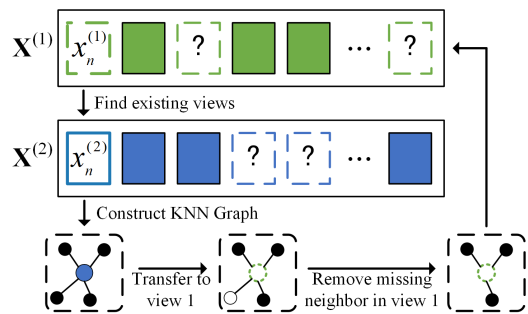}
	\caption{Illustration of constructing graphs on partial multi-view data. Based on the consistency existing in multi-view data, we assume that multi-view data can share similar graphs. Thus the graph of the existing view can be applied to the missing view.}
	\label{fig:graph}
\end{figure}

\paragraph{Constructing initial graph.} For datasets without a predefined graph, graph-based clustering algorithms commonly construct graphs using nearest neighbour algorithm\cite{journals/pami/MalkovY20,conf/icml/GuoSLGSCK20}. However, it is difficult to construct graphs on partial multi-view data because the missing data cannot be used to calculate distance, resulting in incomplete graphs. In this work, cross-view relation graph transfer is devised to construct relation graphs on missing view, which transfers the known inter-instance relation graph in the existing views to recover the missing view graphs. Here we use the two-view data shown in Fig.~\ref{fig:graph} to show the process of constructing graphs on partial multi-view data. For missing data $x^{(1)}_n$ in the first view, the corresponding existing sample in the second view can be denoted as $x^{(2)}_n$. The constructed nearest neighbour graph of $x^{(2)}_n$ can be denoted by $P^{(2)}_n = \{x^{(2)}_{n^1},...,x^{(2)}_{n^K}\}$, where $U_{n^1,2} = ... = U_{n^K,2} = 1$ and $K$ is the number of neighbours. Considering the consistency of multiple views, similarity relationships between samples in existing views are valid for the missing views. Then we can transfer the nearest neighbour graph to the first view to find similar samples to the missing data $x^{(1)}_n$. Since some samples in $P^{(2)}_n$ are missing in the first view, we remove the missing samples and obtain the transferred graph $P^{(1)}_n = \{x^{(1)}_{n^1},...,x^{(1)}_{n^{K'}}\}$, where $K'\leq K$, $U_{n^1,1} = ... = U_{n^{K'},1} = 1$. In this way, we can obtain the graph of all missing data. For datasets with more than two views, there may be more known views for missing data $x^{(1)}_n$. We merge graphs of these views into a single graph to obtain the transferred graph.

\begin{figure*}[t]
	\centering
	\includegraphics[width=16cm]{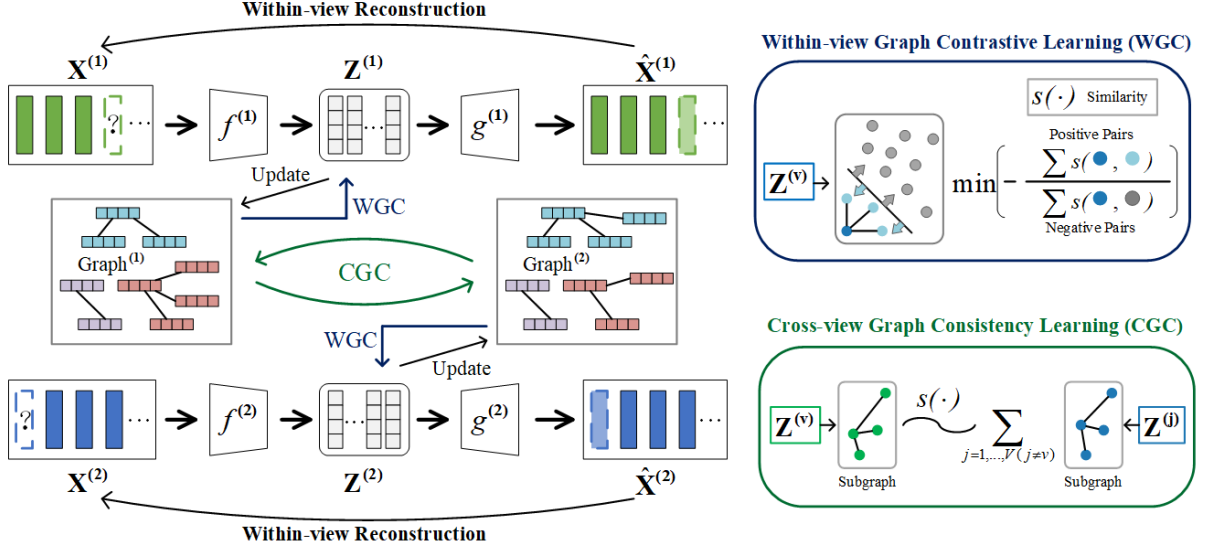}
	\caption{The framework of ACTIVE. In the figure, two-view data is used as a showcase. As shown, it consists of three joint learning objectives: within-view reconstruction, within-view graph contrastive learning (WGC), and cross-view graph consistency learning (CGC). }
	\label{fig:model}
\end{figure*}

\section{ACTIVE: The Proposed Model}

In this section, we present the augmentation-free graph contrastive learning framework in detail. As illustrated in Fig.~\ref{fig:model}, ACTIVE is a joint training framework including three main objectives: within-view reconstruction, within-view graph contrastive learning, and cross-view graph consistency learning. Formally, the objective function of ACTIVE is
\begin{equation}\small
    \mathcal{L} = L_{REC} + \alpha L_{WGC} + \beta L_{CGC}
\end{equation}
where $L_{REC}$, $L_{WGC}$, and $L_{CGC}$ are reconstruction loss, within-view graph contrastive loss, and cross-view graph consistency loss, respectively. $\alpha$ and $\beta$ are the trade-off hyperparameters for within-view graph contrast and cross-view graph consistency, respectively. These three sections are described in more detail below.

As can be seen from Fig.~\ref{fig:model}, the basic structure of ACTIVE is a set of autoencoders. Since the data from different views generally have varying dimensionality, we design view-specific autoencoders to learn representations for each view. Let $f^{(v)}(x^{(v)}_i;\theta^{(v)}_f)$ and $g^{(v)}(z^{(v)}_i;\theta^{(v)}_g)$ denote view-specific encoders and decoders, where $\theta^{(v)}_f$ and $\theta^{(v)}_g$ are network parameters, respectively. $z^{(v)}_i = f^{(v)}(x^{(v)}_i) \in \mathbb{R}^C$ denotes the learned latent representations of $v$-th view with the dimension of $C$ (same as the number of clusters) and $\hat{x}^{(v)}_i = g^{(v)}(z^{(v)}_i) \in \mathbb{R}^{d_v}$ denotes the reconstructed samples of $v$-th view.

The reconstruction loss $Lr^{(v)}_{i}$ between the input $x^{(v)}_i$ and the reconstruction $\hat{x}^{(v)}_i$ is optimized to map the data into latent representations, which can be defined as
\begin{equation}\small
Lr^{(v)}_{i}=\left\{\begin{matrix}
 ||\hat{x}_i^{(v)}- x^{(v)}_i||^2_2, &\ U_{i,v} = 1  \\
 \frac{1}{K}\sum_{k=1}^K ||\hat{x}_i^{(v)}- x^{(v)}_{i^k}||^2_2, & \ U_{i,v} = 0  \\
\end{matrix}\right.
\end{equation}
where $x^{(v)}_{i^k}$ is in the transferred relation graph $P^{(v)}_i$. For existing samples, the decoders rightfully reconstruct them to the samples themselves. For missing samples, the encoders reconstruct them as samples in the transferred graphs to learn common characteristics in the cluster. Then we can obtain the total reconstruction loss as follows.
\begin{equation}\small
L_{REC} = \frac{1}{NV}\sum_{i=1}^N\sum_{v=1}^V Lr^{(v)}_{i}
\end{equation}

\subsection{Within-view Graph Contrastive Learning}

To learn a representation suitable for clustering, we design within-view graph contrastive learning (WGC) to implement cluster-level contrastive learning on multi-view data. Contrastive learning seeks to maximize the similarity of positive pairs while minimizing the similarity of negative pairs. In this work, positive and negative pairs are constructed at the cluster-level according to pseudo-labels generated by the relation graphs. In detail, if one sample is in the relation graph of another sample, they form positive pair sharing the same pseudo-label, otherwise the opposite.

In the training process, the constructed graph of representation $z^{(v)}_i$ can be denoted by $Q^{(v)}_i = \{z^{(v)}_{i^1},...,z^{(v)}_{i^K}\}$. Thus, for each representation $z^{(v)}_i$, $K$ positive pairs can be obtained based on whether a representation is in $Q^{(v)}_i$. Given a mini-batch of size $M$, there are $KM$ pairs in total, of which $K(M-1)$ pairs are negative pairs other than positive pairs.

Then, we adopt cosine distance to compute the similarity between two representations, which is defined as: 
\begin{equation}\small
    s(z^{(v)}_i, z^{(v)}_j) = \frac{(z^{(v)}_i)(z^{(v)}_j)^T}{\| z^{(v)}_i \| \| z^{(v)}_j \|}.
\end{equation}
According to the above discussions, $z^{(v)}_j$ should be similar to $z^{(v)}_i$ if in its relation graph ($P^{(v)}_i$ or $Q^{(v)}_i$) while be far away if not. Therefore, the loss for $z^{(v)}_i$ is defined as
\begin{equation}\small
\begin{aligned}
     & Lw_i^{(v)} = \\
     & -\frac{1}{K}\sum_{k=1}^K \log\frac{\exp{(s(z^{(v)}_i, z^{(v)}_{i^k}))}}{\sum_{m=1}^M\left [\exp{(s(z^{(v)}_i, z^{(v)}_m))} + \exp{(s(z^{(v)}_i, z^{(v)}_{m^k}))}\right ]}, \\
\end{aligned}
\end{equation}
where $z^{(v)}_{i^k}$ is in the relation graph of $z^{(v)}_i$, $z^{(v)}_m$ denotes the representation in the same mini-batch with $z^{(v)}_i$, and $z^{(v)}_{m^k}$ is in the relation graph of $z^{(v)}_m$. In this way, similar samples are more clustered in the latent space, while dissimilar samples maintain a large gap. Then the graph contrastive loss is generalized from the individual to the whole, and the final loss can be described as 
\begin{equation}\small
    L_{WGC} = \frac{1}{NV} \sum_i^N \sum_v^V Lw_i^{(v)} 
\end{equation}

\subsection{Cross-view Graph Consistency Learning}


Some current work \cite{conf/cvpr/0001GLL0021,conf/nips/JiangXYCH19} exploits inter-view consistency to infer missing data by inter-view mapping, making the model more complex, and the mapper may be challenging to train. Differently, we turn to learn a joint semantic space via a low-dimensional representation. More specifically, since different modalities of the same sample share the same high-dimensional semantics, the learned view-specific representations $\{z^{(v)}_i\}_{v=1}^V$ is expected to be similar.

However, instance-level similarity learning is sensitive to the missing data and does not take into account the requirements of the clustering task. We extend it to the graph to better learn consistent representations between views via consistent learning of relation graphs from different views. Then, the inter-view graph-consistency for $z^{(v)}_i$ can be expressed as follows:
\begin{equation}\small
    Lc_i^{(v)} = \frac{1}{K}\sum_{j \neq v}^V\sum_{k=1}^K\|z^{(v)}_{i^k} - z^{(j)}_{i^k}\|_2^2 \label{equ:cgcs}
\end{equation}
where $z^{(v)}_{i^k}$ is in the relation graph of $z^{(v)}_i$. Extending Eq.~\ref{equ:cgcs} from $z^{(v)}_i$ to the whole we can obtain the cross-view graph consistency loss as follows.
\begin{equation}\small
    L_{CGC} = \frac{1}{NV} \sum_i^N \sum_v^V Lc_i^{(v)} 
\end{equation}

\subsection{Implementation}

As mentioned above, nearest neighbour graphs $P^{(v)}_i$ and $Q^{(v)}_i$ are constructed based on the distance of sample $x^{(v)}_i$ and representation $z^{(v)}_i$, respectively. We divide the training process of the model into two stages according to the graph used in WGC and CGC.

\paragraph{Stage {\romannumeral1}: Training via the initial graph $P^{(v)}_i$.}
At the early phases of training, the network cannot learn the feasible features of the original data. Thus, $P^{(v)}_i$ constructed using the original features is used as the graph to compute $L_{WGC}$ and $L_{CGC}$. However, it is unavoidable that some samples from different clusters are linked together in the initial graph, which may worsen the learning of representations and limit clustering performance. Then, after a few hundred epochs, training of the model goes to stage {\romannumeral2}.

\paragraph{Stage {\romannumeral2}: Training via the learned graph $Q^{(v)}_i$.} In the second training stage, $Q^{(v)}_i$ is constructed to compute $L_{WGC}$ and $L_{CGC}$. As training proceeds, the learned representation is increasingly discriminative, thus improves the quality of the graph $Q^{(v)}_i$, while the high quality graph in turn enhances the separability of the latent representation.

Since representations of different views for the same sample are expected to be similar in training, we sum them directly to obtain the common representation $z_i^*$. After the two training stages, following the previous work on deep clustering \cite{conf/ijcai/XuGZWNL19,conf/mm/0001ZZWFXZ20}, we fine-tune the network with KL divergence-based clustering loss and obtain the final clustering results. More details on the clustering fine-tuning can be found in the supplementary material.

\section{Experiments}

In this section, we describe the experiments conducted to demonstrate the effectiveness of the proposed ACTIVE. The experiments aim to answer the following research questions: 
\textbf{(Q1)} How does ACTIVE perform in partial multi-view clustering? 
\textbf{(Q2)} How does each component influence the performance of ACTIVE?
\textbf{(Q3)} Can the model behave stably when changing the hyperparameters?
\textbf{(Q4)} How does the two-stage optimization influence ACTIVE?

\subsection{Experimental Setups}

\paragraph{Baselines:} We consider 10 baselines to demonstrate the effectiveness of ACTIVE, including concat feature K-means++ (Concat) \cite{conf/ijcai/ZhaoLF16}, PVC \cite{conf/aaai/LiJZ14}, MIC \cite{conf/pkdd/ShaoHY15}, IMG \cite{conf/dicta/QianSGTD16}, DAIMC \cite{conf/ijcai/HuC18}, UEAF \cite{conf/aaai/WenZ0ZFL19}, CDIMC-net \cite{conf/ijcai/WenZ0ZFX20}, iCmSC \cite{journals/tip/WangLSGJ21}, GP-MVC \cite{journals/tip/WangDTGF21}, and COMPLETER \cite{conf/cvpr/0001GLL0021}. 

\paragraph{Datasets:} The following five datasets are used to assess the proposed method.
1) \textbf{BDGP} contains 2500 images of five categories, and each image is described by a 1750-D visual vector and a 79-D textual feature vector.
2) \textbf{MNIST} is a widely-used dataset composed of 70000 digital images. Since some baselines cannot handle such a large dataset, we follow \cite{conf/ijcai/WenZ0ZFX20} to use a subset of MNIST, which contains 4000 samples from 10 categories. Pixel feature and edge feature are extracted as two views. 
3) \textbf{Caltech101-20} consists of 2386 images of 20 subjects, and we follow \cite{conf/cvpr/0001GLL0021} to use two views, \emph{i.e.} , HOG and GIST features. 
4) \textbf{Animal} consists of 10,158 images from 50 classes, and two types of deep features are used as two views. 
5) \textbf{Coil20} contains 20 objects and each object provides 72 images with different angles. Three types of features, i.e., LBP, HOG, and gray-value of pixels, are extracted from images as three views.

\paragraph{Partial data construction:} To evaluate the performance of handling partial multi-view data, for datasets with more than two views, we randomly remove $p\%$ samples from every view under the condition that all samples at least have one view. For datasets with two views, $p\%$ samples are randomly selected as paired samples whose views are complete, and the remaining samples are treated as single view samples. 

\paragraph{Evaluation metric:}  Accuracy (ACC), Normalized Mutual Information (NMI), and Adjusted Rand Index (ARI) are used to evaluate the clustering performance. For each metric, a larger value implies a better clustering result. Please refer to the supplementary material for the results of ARI.

\paragraph{Parameters settings:} The common parameters for training ACTIVE are set as $Max\ training\ epoch = 2000$, $Stage\ ii\ start\ epoch = 300$, $Learning\ rate = 0.01$, and $Batch\ size = 128$. Please refer to the supplementary material for the details of network architectures.

\subsection{Performance Comparisons and Analysis (Q1)}

\begin{table*}[ht]
\centering
\resizebox{\textwidth}{!}{
\begin{threeparttable}
\begin{tabular}{c|l|cc|cc|cc|cc|cc}
\toprule
\multirow{2}{*}{$p\%$\tnote{1}} & \multirow{2}{*}{Method}  & \multicolumn{2}{c|}{BDGP} & \multicolumn{2}{c|}{Caltech101-20} & \multicolumn{2}{c|}{MNIST} & \multicolumn{2}{c|}{Animal} & \multicolumn{2}{c}{Coil20} \\
                                                     &                          & ACC  & NMI                & ACC  & NMI                         & ACC  & NMI                        & ACC  & NMI                 & ACC  & NMI             \\
\midrule
\multirow{11}{*}{$p=10$}    
                            & Concat                    & 42.41$\pm$1.73 & 30.11$\pm$4.56 &	32.95$\pm$1.77 & 39.12$\pm$0.95 & 22.35$\pm$0.74 & 16.36$\pm$0.90 & 38.09$\pm$1.44 & 52.04$\pm$0.75 & 54.58$\pm$1.95 & 63.67$\pm$1.05  \\
                            & PVC                       & 57.89$\pm$8.48 & 33.88$\pm$8.05 & 33.82$\pm$2.03 & 49.41$\pm$1.30 & 43.54$\pm$2.02 & 36.74$\pm$0.81 & $-$            & $-$            & 62.00$\pm$1.63 & 74.66$\pm$1.07  \\
                            & MIC                       & 34.74$\pm$3.26 & 16.91$\pm$4.85 & 30.67$\pm$2.60 & 40.43$\pm$1.95 & 34.43$\pm$3.65 & 27.09$\pm$3.36 & 35.35$\pm$0.98 & 49.07$\pm$0.46 & 58.51$\pm$4.12 & 67.99$\pm$1.79  \\
                            & IMG                       & 51.89$\pm$4.17 & 33.70$\pm$3.07 & 38.11$\pm$1.66 & 47.90$\pm$0.97 & 42.55$\pm$2.31 & 32.34$\pm$1.31 & 49.91$\pm$0.11 & 54.16$\pm$0.09 & 60.27$\pm$1.16 & 74.47$\pm$0.69  \\
                            & DAIMC                     & 57.18$\pm$6.02 & 31.53$\pm$5.53 & 32.56$\pm$3.22 & 44.51$\pm$2.71 & 41.07$\pm$3.07 & 33.53$\pm$2.53 & 40.81$\pm$1.34 & 49.00$\pm$0.73 & 71.61$\pm$2.34 & 80.09$\pm$1.43  \\
                            & UEAF                      & 68.79$\pm$2.00 & 42.89$\pm$2.01 & 39.22$\pm$2.27 & 55.78$\pm$1.23 & 45.90$\pm$1.80 & 40.13$\pm$1.05 & 46.08$\pm$1.70 & \textbf{55.48$\pm$0.43} & 68.67$\pm$1.90 & 77.92$\pm$1.06  \\
                            & CDIMC-net                 & 55.04$\pm$4.80 & 36.88$\pm$4.73 & 45.95$\pm$4.72 & 58.32$\pm$1.96 & 49.30$\pm$3.38 & 44.51$\pm$4.30 & 38.10$\pm$2.02 & 52.58$\pm$1.28 & 76.09$\pm$2.56 & 84.68$\pm$0.80  \\
                            & iCmSC                     & 53.04$\pm$2.18 & 39.27$\pm$1.38 & 46.81$\pm$0.59 & 60.05$\pm$0.91 & 46.97$\pm$0.26 & 48.66$\pm$1.35 & 37.40$\pm$1.58 & 52.82$\pm$1.01 & $-$            & $-$             \\
                            & GP-MVC\tnote{2}           & 58.74$\pm$2.49 & $-$            & $-$            & $-$            & \textbf{56.46$\pm$2.47} & $-$            & $-$            & $-$            & $-$            & $-$             \\
                            & COMPLETER                 & 34.24$\pm$6.39 & 18.94$\pm$5.91 & 32.61$\pm$4.39 & 36.97$\pm$7.61 & 47.26$\pm$2.93 & 44.99$\pm$2.49 & 43.26$\pm$3.67 & 55.44$\pm$4.36 & $-$            & $-$             \\
                            & ACTIVE                    & \textbf{81.28$\pm$2.36} & \textbf{59.92$\pm$3.01} & \textbf{64.50$\pm$3.67} & \textbf{61.61$\pm$1.01} & 54.78$\pm$2.62 & \textbf{53.49$\pm$2.21} & \textbf{46.29$\pm$1.98} & 55.37$\pm$0.65 & \textbf{83.98$\pm$1.42} & \textbf{89.78$\pm$1.68}  \\
\midrule
\multirow{11}{*}{$p=30$}    
                            & Concat                    & 46.73$\pm$2.10 & 34.46$\pm$1.99 & 32.48$\pm$1.00 & 44.08$\pm$0.66 & 31.87$\pm$1.31 & 28.00$\pm$1.79 & 38.94$\pm$2.53 & 53.29$\pm$1.00 & 33.42$\pm$1.06 & 45.79$\pm$1.18  \\
                            & PVC                       & 52.43$\pm$4.95 & 30.65$\pm$3.55 & 37.34$\pm$0.80 & 52.31$\pm$0.83 & 47.47$\pm$2.22 & 38.91$\pm$1.13 & $-$            & $-$            & 57.77$\pm$2.59 & 68.13$\pm$1.70  \\
                            & MIC                       & 42.71$\pm$2.18 & 21.31$\pm$2.46 & 28.56$\pm$2.38 & 40.05$\pm$1.19 & 31.29$\pm$2.36 & 26.79$\pm$2.69 & 37.39$\pm$1.26 & 50.97$\pm$0.86 & 38.20$\pm$1.63 & 50.29$\pm$2.33  \\
                            & IMG                       & 52.39$\pm$1.48 & 35.76$\pm$1.77 & 42.44$\pm$1.27 & 53.36$\pm$1.02 & 42.77$\pm$1.76 & 33.84$\pm$1.17 & 54.52$\pm$0.23 & 58.58$\pm$0.09 & 57.14$\pm$0.81 & 68.76$\pm$0.69  \\
                            & DAIMC                     & 57.57$\pm$5.84 & 34.91$\pm$4.36 & 41.72$\pm$3.19 & 55.14$\pm$2.08 & 47.68$\pm$3.33 & 40.80$\pm$2.93 & 51.18$\pm$1.59 & 55.87$\pm$0.68 & 70.80$\pm$1.87 & 78.67$\pm$1.03  \\
                            & UEAF                      & 77.50$\pm$6.35 & 55.23$\pm$5.29 & 45.62$\pm$2.39 & 59.50$\pm$1.21 & 47.28$\pm$2.10 & 42.48$\pm$1.20 & 53.62$\pm$1.00 & 60.58$\pm$0.47 & 67.46$\pm$2.37 & 76.33$\pm$1.18  \\
                            & CDIMC-net                 & 76.25$\pm$7.63 & 59.04$\pm$8.68 & 47.50$\pm$5.62 & 60.09$\pm$2.15 & 53.23$\pm$4.53 & 49.11$\pm$5.16 & 44.16$\pm$1.42 & 55.37$\pm$0.61 & 75.18$\pm$2.21 & 83.28$\pm$0.91  \\
                            & iCmSC                     & 67.90$\pm$2.09 & 44.63$\pm$2.64 & 45.88$\pm$1.17 & 59.49$\pm$0.47 & 51.11$\pm$0.58 & 53.27$\pm$0.40 & 40.27$\pm$1.01 & 54.74$\pm$1.51 & $-$            & $-$             \\
                            & GP-MVC                    & 78.68$\pm$2.34 & $-$            & $-$            & $-$            & 55.42$\pm$3.30 & $-$            & $-$            & $-$            & $-$            & $-$             \\
                            & COMPLETER                 & 50.75$\pm$5.68 & 37.90$\pm$4.31 & 66.68$\pm$4.42 & 63.44$\pm$1.92 & 50.04$\pm$2.72 & 45.80$\pm$3.20 & 44.45$\pm$2.59 & 57.45$\pm$2.59 & $-$            & $-$             \\
                            & ACTIVE                    & \textbf{91.36$\pm$1.19} & \textbf{76.06$\pm$2.00} & \textbf{70.49$\pm$3.03} & \textbf{66.44$\pm$0.69} & \textbf{57.33$\pm$1.85} & \textbf{54.54$\pm$1.21} & \textbf{55.02$\pm$1.43} & \textbf{61.10$\pm$0.44} & \textbf{83.13$\pm$0.81} & \textbf{88.83$\pm$0.90}  \\
\midrule
\multirow{11}{*}{$p=50$}    
                            & Concat                    & 51.26$\pm$2.58 & 39.13$\pm$2.58 & 34.51$\pm$1.92 & 49.28$\pm$1.23 & 33.64$\pm$1.65 & 27.73$\pm$1.48 & 42.54$\pm$1.03 & 56.34$\pm$0.59 & 26.64$\pm$1.22 & 41.17$\pm$1.38  \\
                            & PVC                       & 53.87$\pm$5.73 & 32.32$\pm$4.55 & 39.57$\pm$1.42 & 53.58$\pm$1.33 & 45.81$\pm$2.91 & 39.53$\pm$1.38 & $-$            & $-$            & 50.49$\pm$0.99 & 58.42$\pm$0.89  \\
                            & MIC                       & 53.84$\pm$2.16 & 31.59$\pm$1.94 & 32.94$\pm$2.26 & 46.17$\pm$0.95 & 34.83$\pm$1.66 & 29.22$\pm$1.66 & 41.86$\pm$1.42 & 55.25$\pm$0.73 & 25.10$\pm$2.41 & 37.22$\pm$2.72  \\
                            & IMG                       & 51.00$\pm$1.84 & 36.97$\pm$1.45 & 43.63$\pm$1.97 & 55.46$\pm$1.39 & 46.22$\pm$3.12 & 36.49$\pm$1.44 & 56.81$\pm$0.19 & 61.40$\pm$0.10 & 51.08$\pm$1.13 & 60.45$\pm$0.89  \\
                            & DAIMC                     & 62.02$\pm$4.48 & 43.36$\pm$2.99 & 45.78$\pm$2.43 & 60.86$\pm$1.05 & 49.40$\pm$3.02 & 43.18$\pm$2.17 & 54.42$\pm$1.22 & 60.30$\pm$0.48 & 66.26$\pm$1.89 & 73.75$\pm$1.95  \\
                            & UEAF                      & 83.34$\pm$8.56 & 64.35$\pm$5.56 & 46.77$\pm$2.49 & 60.77$\pm$1.07 & 49.85$\pm$1.19 & 44.63$\pm$0.77 & 57.89$\pm$1.12 & 64.66$\pm$0.41 & 60.41$\pm$8.59 & 68.18$\pm$5.98  \\
                            & CDIMC-net                 & 83.40$\pm$7.01 & 67.46$\pm$8.43 & 51.71$\pm$3.65 & 64.08$\pm$1.14 & 55.46$\pm$3.77 & 52.12$\pm$5.04 & 49.64$\pm$2.06 & 61.48$\pm$1.02 & 72.52$\pm$2.73 & 81.33$\pm$1.24  \\
                            & iCmSC                     & 80.27$\pm$0.74 & 61.28$\pm$1.32 & 51.90$\pm$1.68 & 61.80$\pm$1.59 & 56.05$\pm$1.12 & 54.21$\pm$1.20 & 44.31$\pm$4.00 & 55.73$\pm$0.70 & $-$            & $-$             \\
                            & GP-MVC                    & 88.79$\pm$1.28 & $-$            & $-$            & $-$            & 57.76$\pm$1.69 & $-$            & $-$            & $-$            & $-$            & $-$             \\
                            & COMPLETER                 & 51.92$\pm$5.95 & 38.07$\pm$3.52 & 70.18$\pm$4.72 & 69.58$\pm$2.09 & 52.56$\pm$2.33 & 49.76$\pm$2.25 & 52.36$\pm$3.60 & 61.15$\pm$3.50 & $-$            & $-$             \\
                            & ACTIVE                    & \textbf{93.16$\pm$3.45} & \textbf{80.67$\pm$3.51} & \textbf{71.63$\pm$1.83} & \textbf{70.90$\pm$2.04} & \textbf{58.10$\pm$2.34} & \textbf{54.82$\pm$1.20} & \textbf{61.91$\pm$1.19} & \textbf{68.99$\pm$0.90} & \textbf{78.46$\pm$1.20} & \textbf{84.21$\pm$0.63}  \\
\bottomrule
\end{tabular}
\begin{tablenotes}
\item[1] The results on these datasets with a paired-view rate of $p=70$ can be find in the supplementary material.
\item[2] The results of GP-MVC are reported in \cite{journals/tip/WangDTGF21}.
\end{tablenotes}
\end{threeparttable}
}
\caption{Clustering average results and standard deviations on five datasets with different missing-view rates or paired-view rates $p\%$}\label{tab:comparsion}
\vspace{-0.5em}
\end{table*}

\begin{figure*}[ht]
\small
\centering
\subfigure[IMG (52.4\%)]{\includegraphics[width=3.4cm]{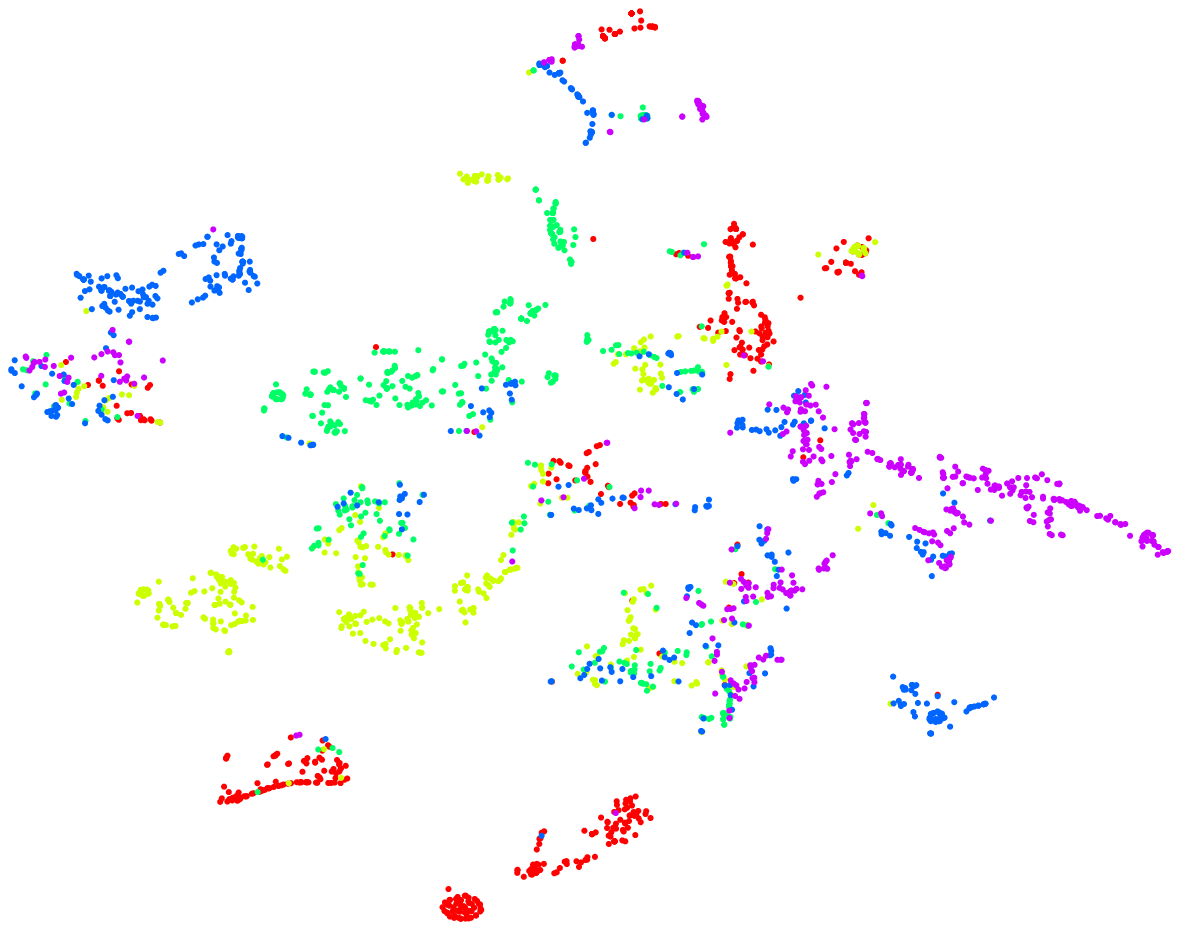}}\vspace{-0.2em}
\subfigure[DAIMC (57.6\%)]{\includegraphics[width=3.4cm]{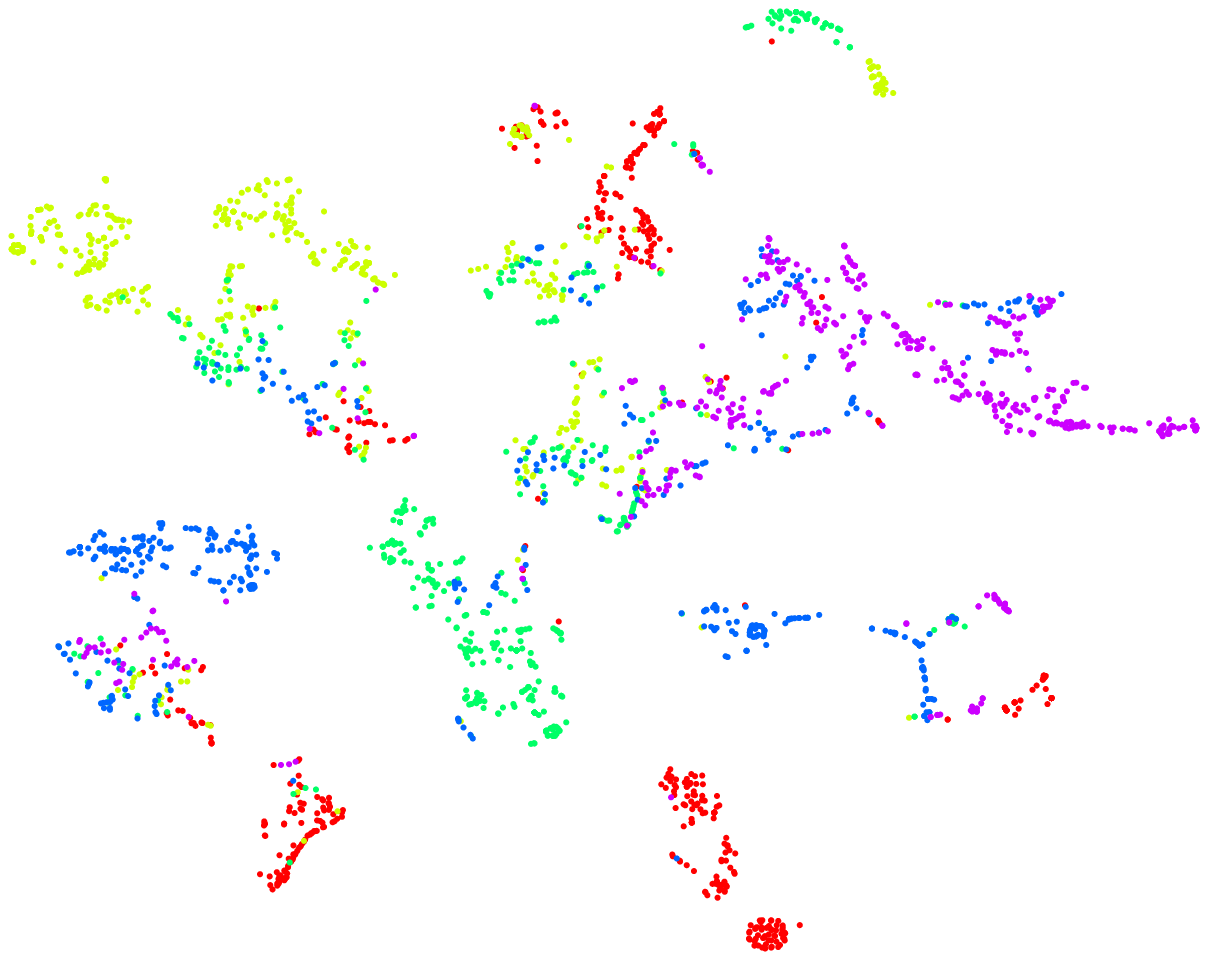}}\vspace{-0.2em}
\subfigure[CDIMC-net (76.3\%)]{\includegraphics[width=3.4cm]{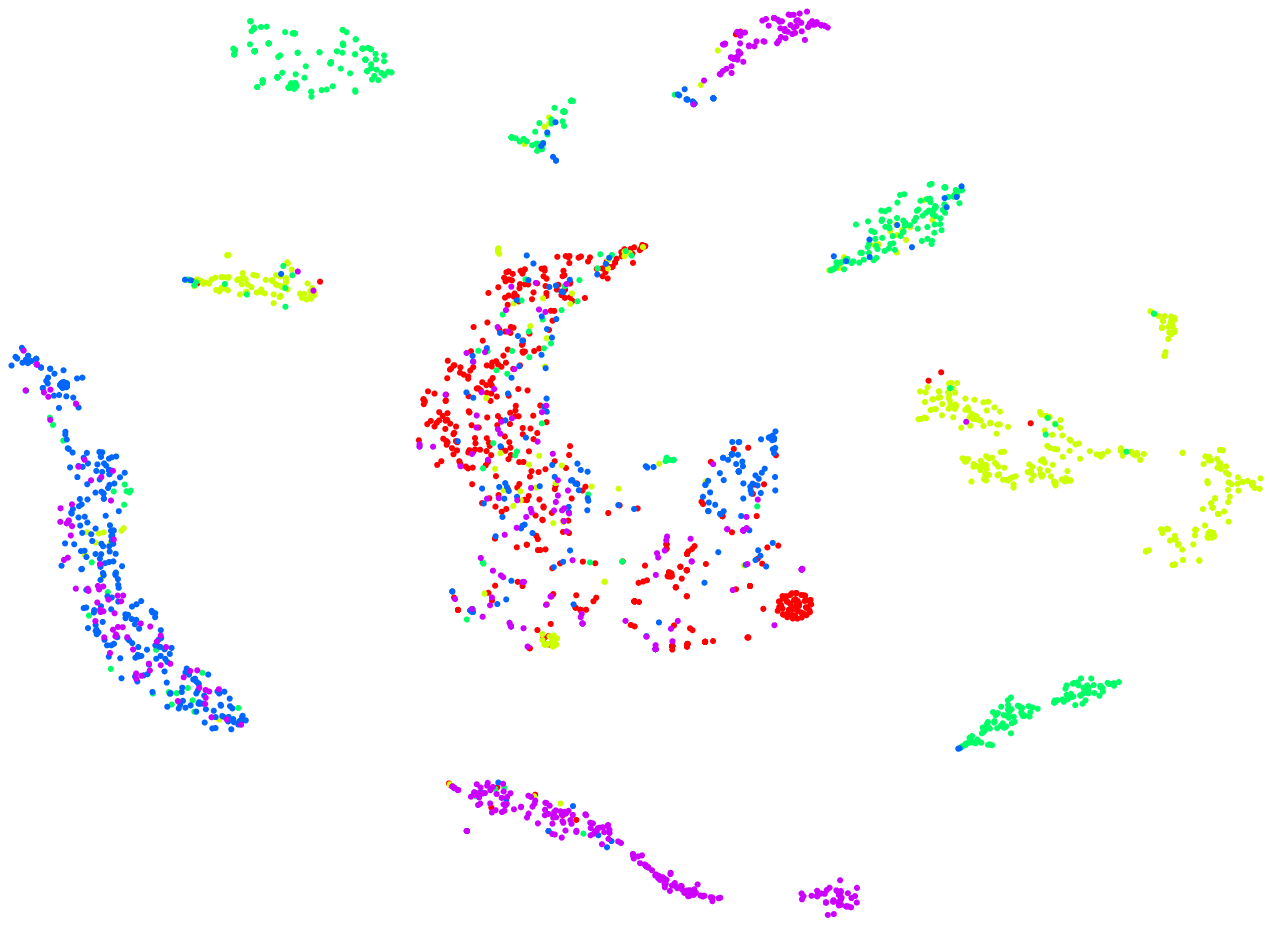}}\vspace{-0.2em}
\subfigure[COMPLETER (50.4\%)]{\includegraphics[width=3.5cm]{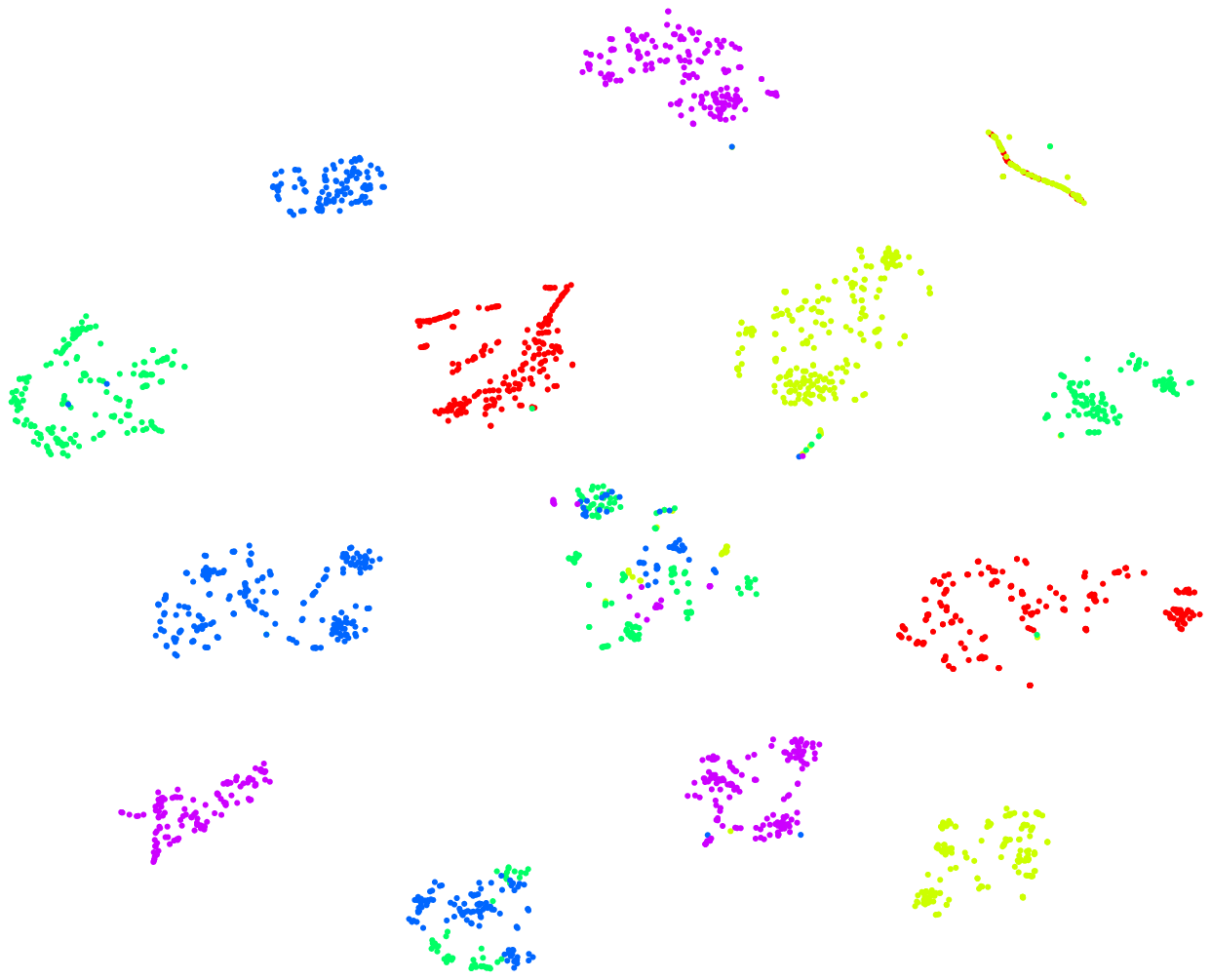}}\vspace{-0.2em}
\subfigure[ACTIVE (91.4\%)]{\includegraphics[width=3.4cm]{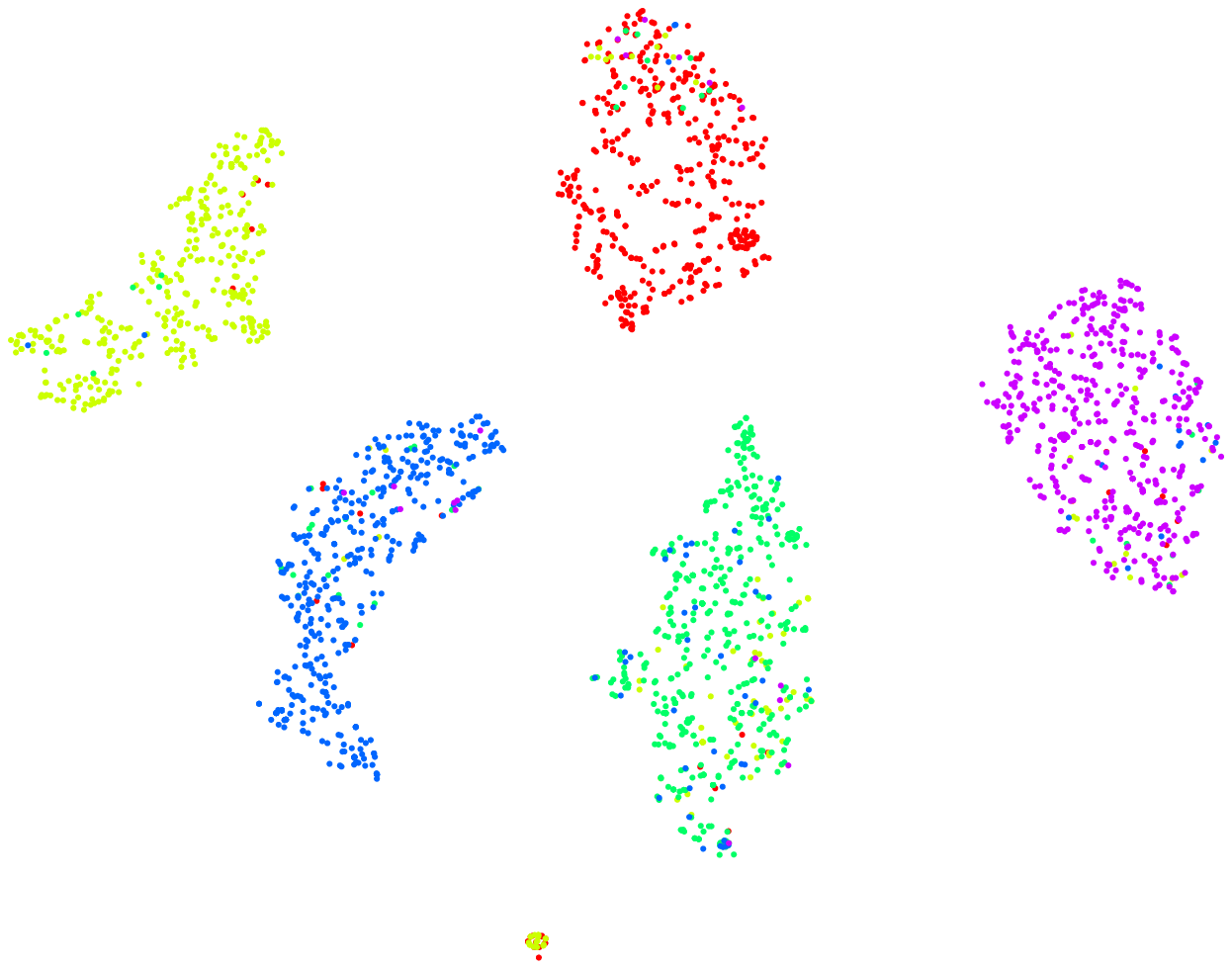}}
\caption{$t$-SNE visualizations of representations learned by various methods on BDGP with the paired-view rates of 30\%, where different colours indicate different categories. The average ACC (\%) is marked in the subplots.}
\label{fig2}\label{fig:vis}
\vspace{-0.5em}
\end{figure*}

The experimental results of different PMVC methods on the five datasets are given in Table.~\ref{tab:comparsion}, where the best results are highlighted in \textbf{bold}. $``-"$ indicates unavailable results because the method cannot handle data with more than two views or out of memory. From the experimental results, we have the following observations: (1) Compared with all baselines, the proposed ACTIVE achieves the best performance on all datasets with most evaluation metrics. Especially on the BDGP dataset with a missing-view rate of $30\%$, ACTIVE obtains about $91\%$ ACC and $76\%$ NMI, which are about $12\%$ and $17\%$ higher than those of the second-best method, respectively. (2) ACTIVE consistently outperforms the advanced GAN-based approach GP-MVC and the contrastive learning-based approach COMPLETER, which demonstrates the effectiveness of ACTIVE. (3) Almost all methods achieve better performance than Concat in most cases, showing that exploiting the complementarity and consistency of multi-view data can improve clustering performance.

To more intuitively show the superiority of the representation obtained by our ACTIVE, $t$-SNE \cite{jmlr/laurens08} is used to visualize the results of different methods on the BDGP dataset with a paired-view ratio of 30\% in Fig.~\ref{fig:vis}. As can be seen from the figure, our method can recover a better cluster structure of data since it has smaller intra-cluster scatter and larger inter-cluster scatter, which demonstrates the effectiveness of the proposed ACTIVE.

\begin{figure}[t]
\centering
\subfigure[Imputation comparison]
{
\label{subfig:imput}
\includegraphics[width=4cm]{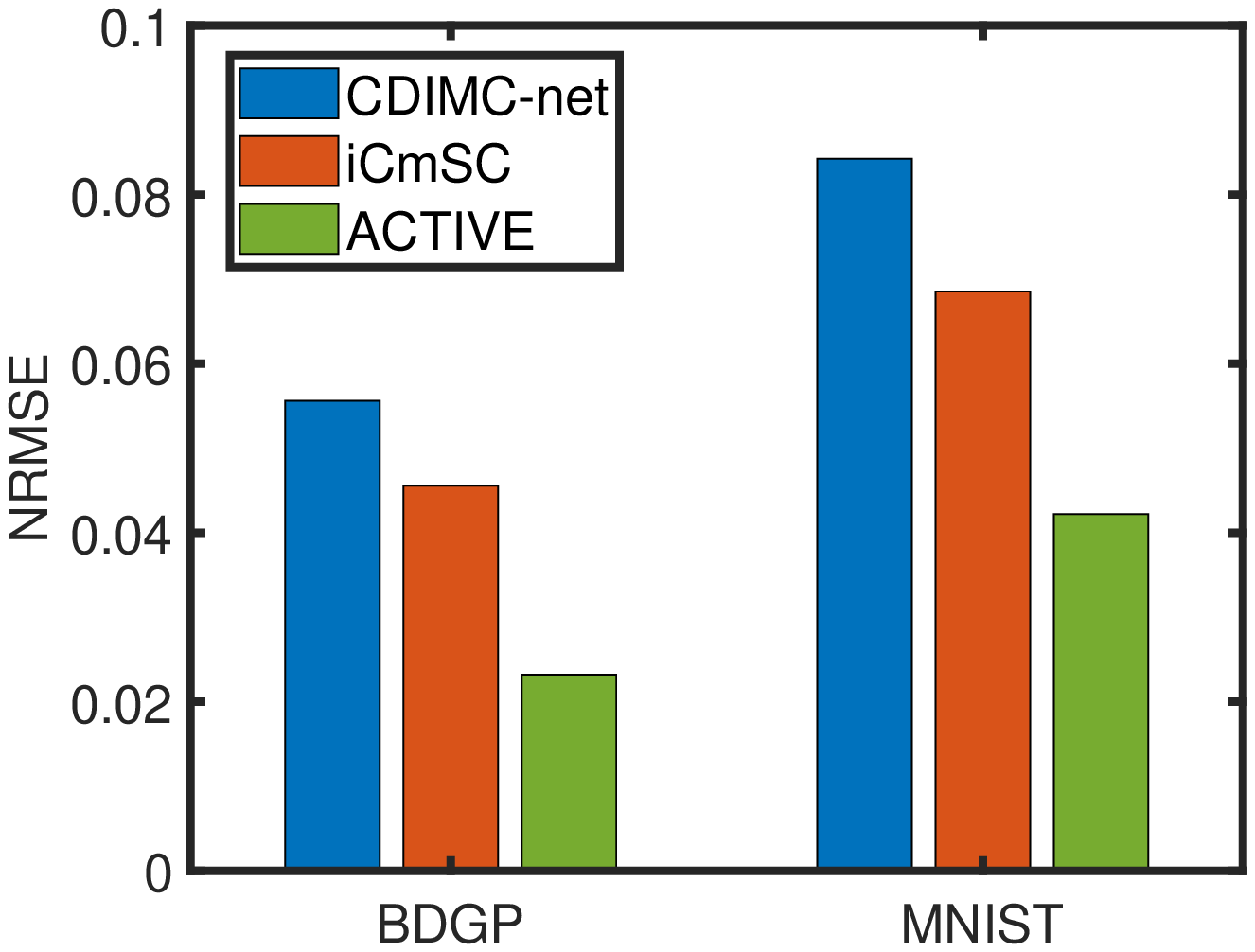}
}
\subfigure[$K$-sensitivity analysis]
{
\label{subfig:k}
\includegraphics[width=4cm]{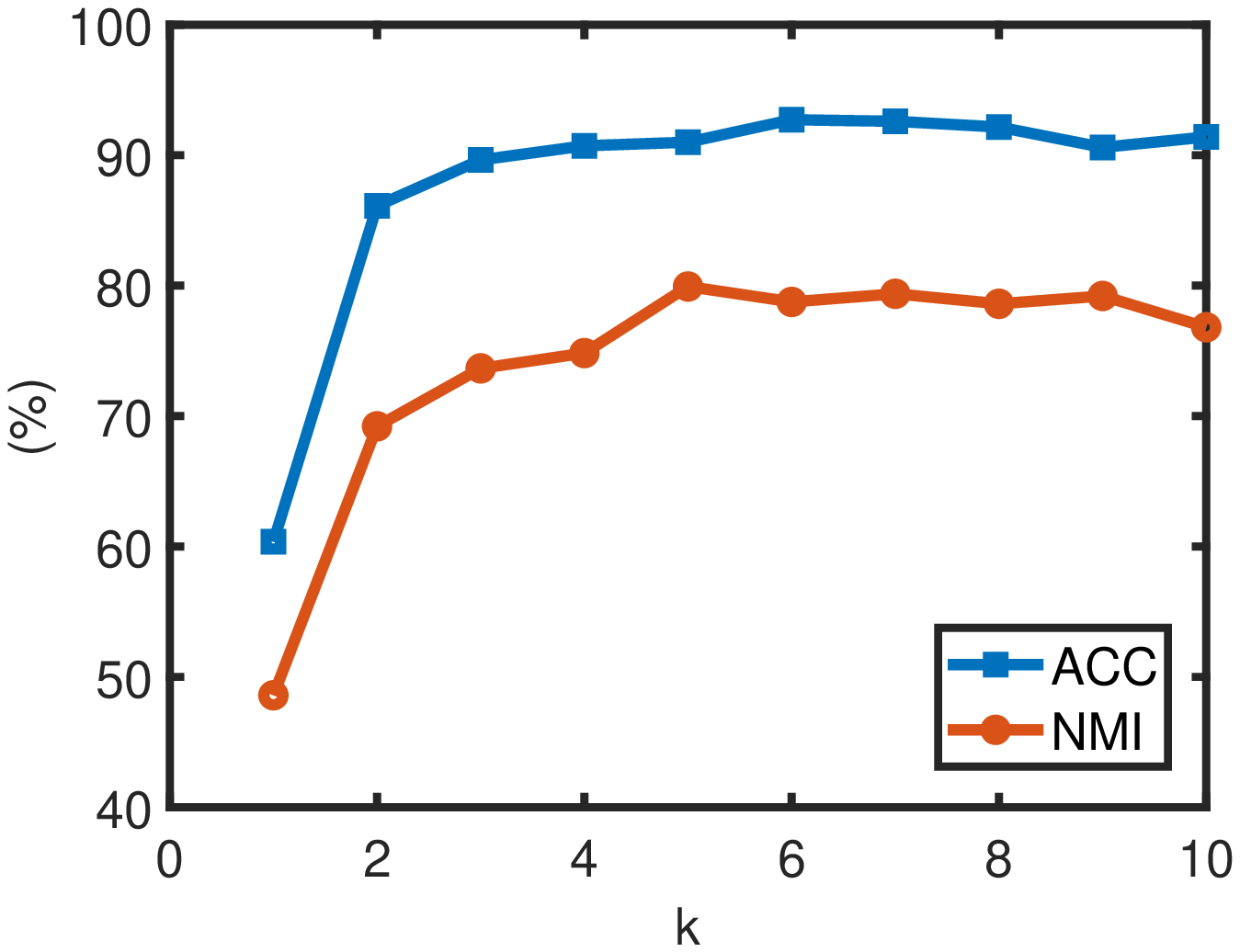}
}
\vfill
\subfigure[Parameter analysis on Coil20 with a missing-view rate of $30\%$]
{
\label{subfig:para}
\includegraphics[width=4.3cm]{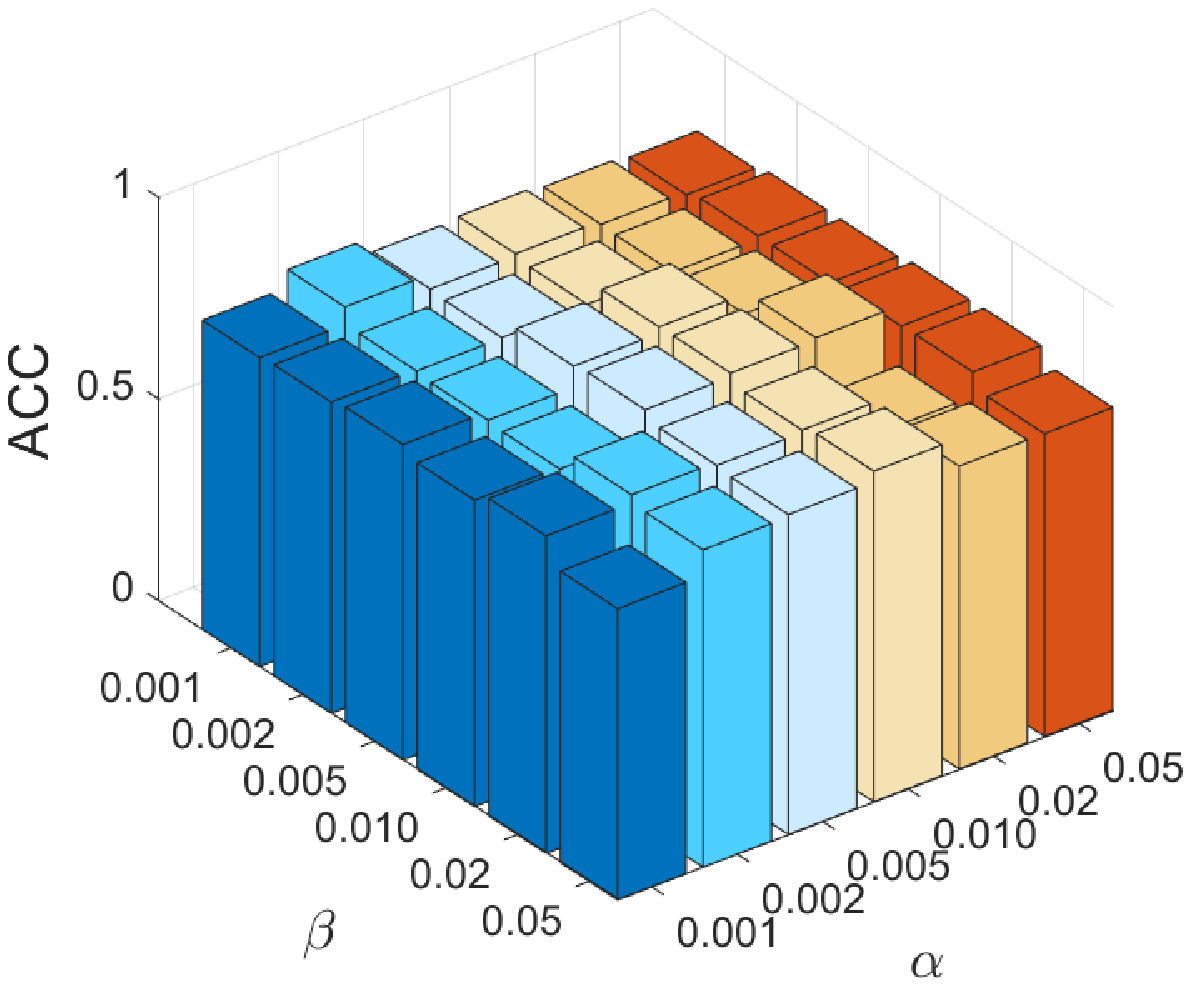}\vspace{-1em}
\includegraphics[width=4.3cm]{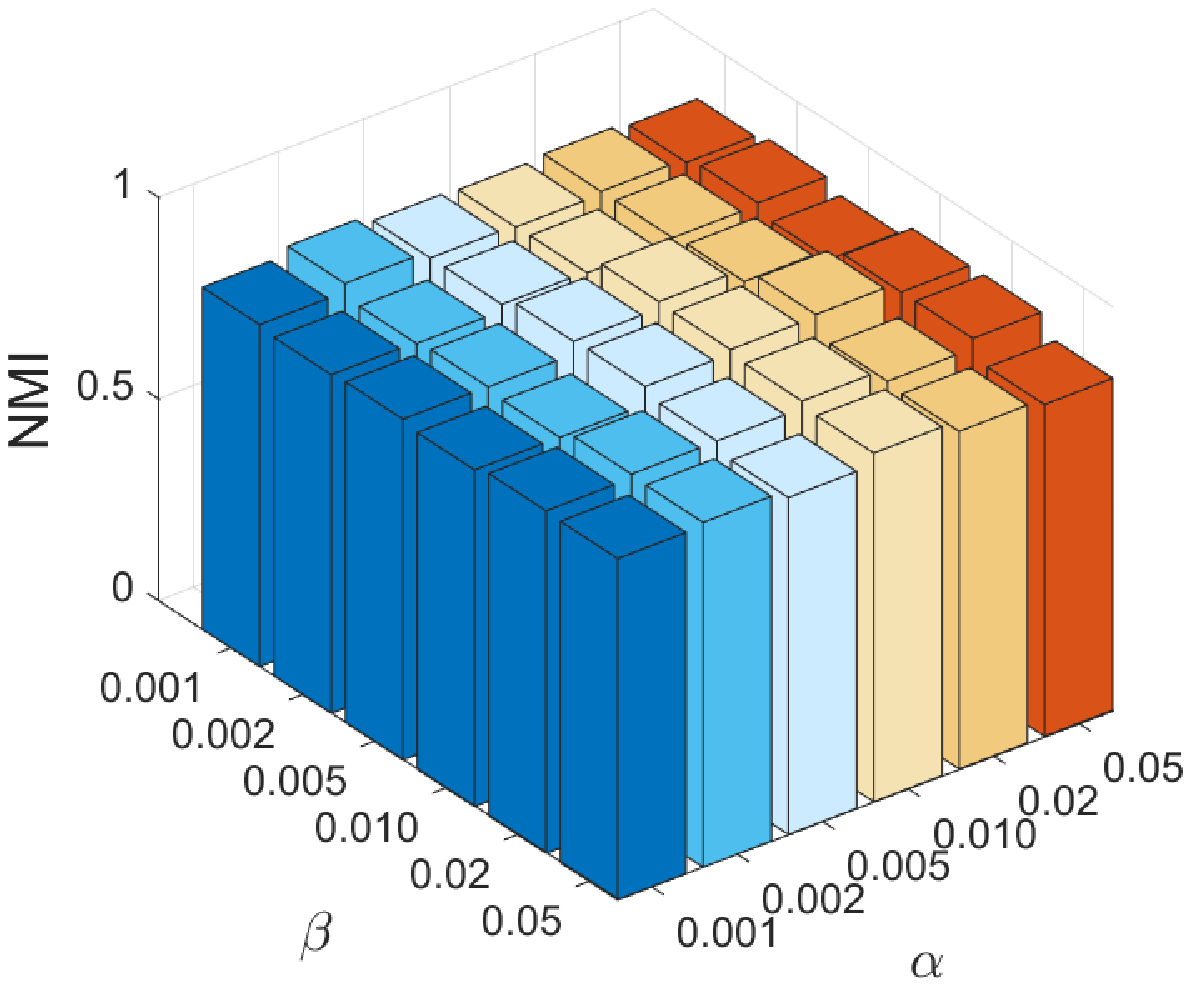}
}\vspace{-0.5em}

\vfill
\subfigure[Influence of training stage {\romannumeral2} starting epochs]
{
\label{subfig:stage2}
\includegraphics[width=4cm]{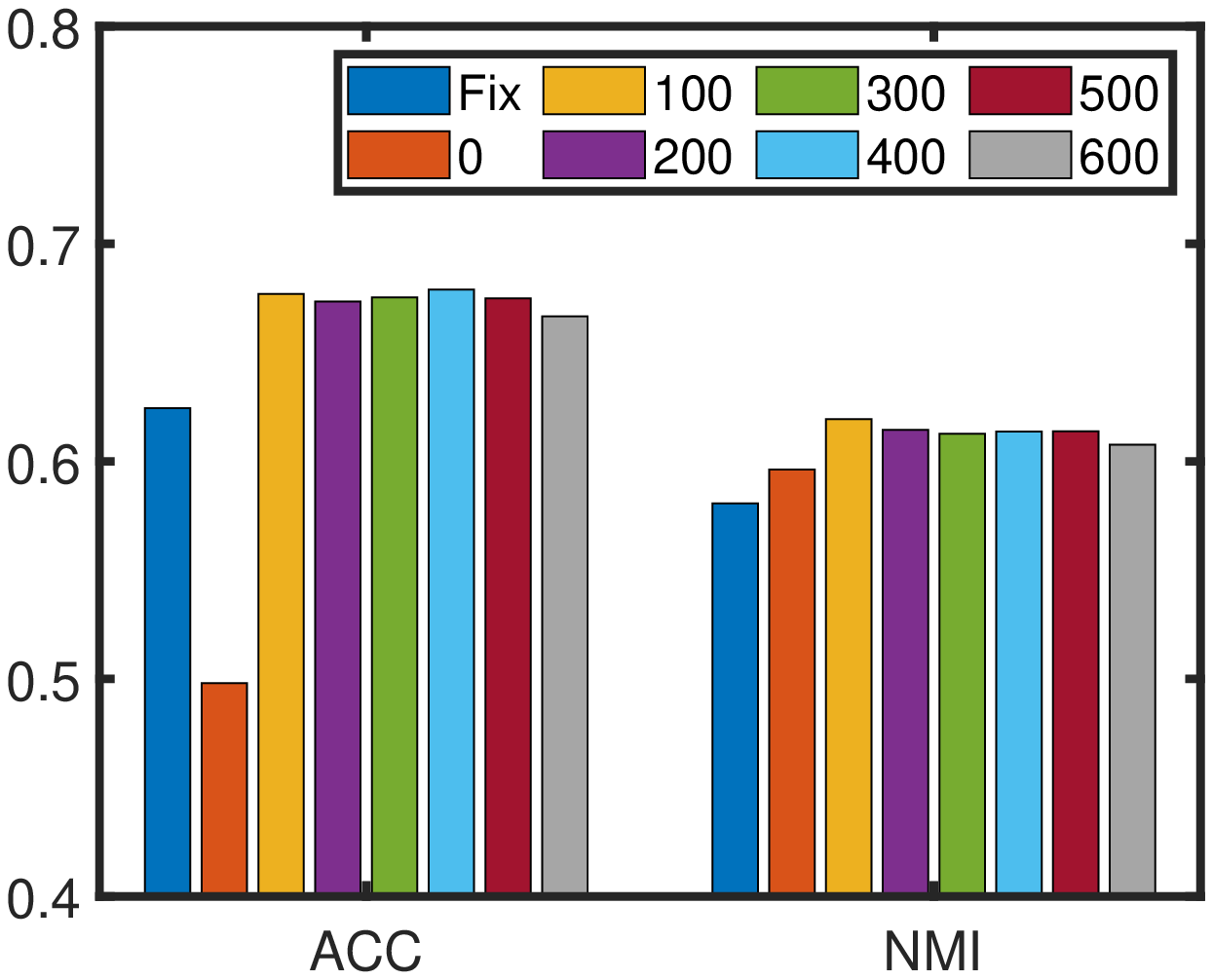}
}
\subfigure[Loss and the error of the graph v.s. epochs]
{
\label{subfig:train}
\includegraphics[width=4cm]{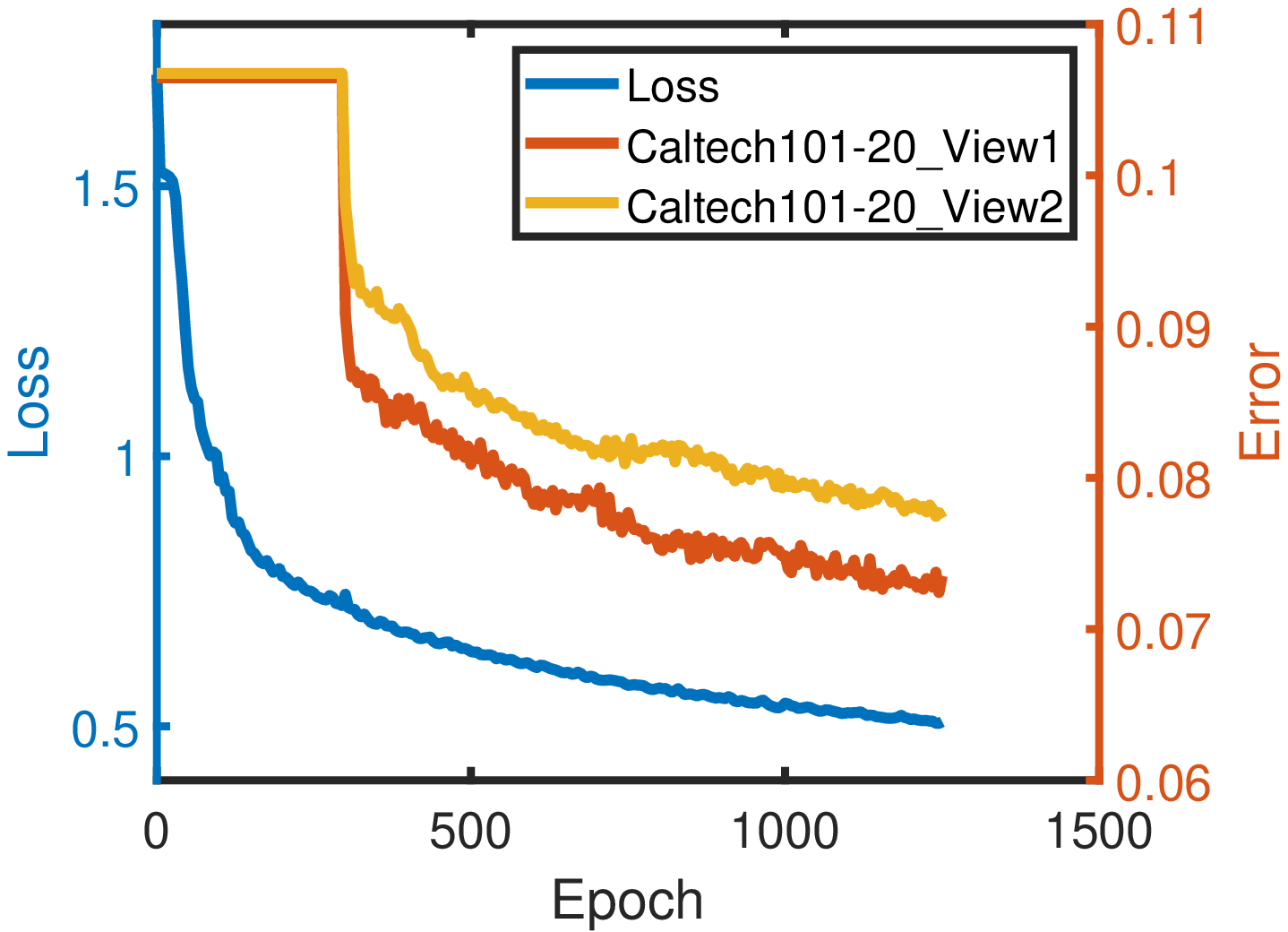}
}
\caption{Imputation comparison and parameter/training analyses}
\label{fig:analy}
\vspace{-0.5em}
\end{figure}

Unlike most existing partial multi-view clustering methods, in our ACTIVE, the constructed inter-instance relation graphs are transferred to the missing views to impute missingness. We evaluate the imputation performance of ACTIVE in terms of Normalized Root Mean Square Error (NRMSE), where a lower value indicates a better performance. The imputation performance on BDGP and MNIST (with a missing-view rate of $30\%$) is shown in Fig.~\ref{subfig:imput}. As shown in the figure, ACTIVE outperforms the other two deep learning-based approaches on these two datasets, illustrating the superior imputation performance of ACTIVE.

\subsection{Ablation Study (Q2)}

To verify the importance of each component in ACTIVE, we perform ablation studies to isolate the effect of the reconstruction loss $L_{REC}$, within-view graph contrastive loss $L_{WGC}$, and cross-view graph consistency loss $L_{CGC}$. From the results in Table.~\ref{tab:abl}, one could observe that: (1) The best performance can be achieved when using whole loss terms, indicating that all the components play indispensable roles in ACTIVE. (2) The reconstruction loss plays an essential role in the AEs, nevertheless adding any loss of $L_{WGC}$ and $L_{CGC}$ on this base can boost the performance. This suggests that WGC and CGC both contribute to relieving the impact of absence and learning cluster-friendly representations.

\begin{table}[t]
\centering
\resizebox{7cm}{!}{
\begin{tabular}{cccccc}
\toprule
$L_R$ & $L_{WGC}$ & $L_{CGC}$ & ACC    & NMI    & ARI   \\
\midrule
\Checkmark         &                   &                  & 56.75   & 66.45  & 47.01  \\
                   & \Checkmark        &                  & 42.25   & 53.02  & 26.45  \\
                   &                   & \Checkmark       & 33.62   & 26.11  & 11.13  \\
                   & \Checkmark        & \Checkmark       & 39.74   & 24.98  & 12.29  \\
\Checkmark         &                   & \Checkmark       & 68.09   & 68.98  & 61.56  \\
\Checkmark         & \Checkmark        &                  & 61.57   & 68.80  & 56.02  \\   
\Checkmark         & \Checkmark        & \Checkmark       & \textbf{71.63}   & \textbf{79.90}  & \textbf{66.44}  \\
\bottomrule
\end{tabular}}
\caption{Ablation study on Caltech101-20 dataset with a paired-view rates of 50\%. Different methods use modules identified by \Checkmark.}\label{tab:abl}
\end{table}

\subsection{Parameter Analysis (Q3)}

The number of the nearest neighbours $K$ is an important parameter in constructing the nearest neighbour graph and has a great impact on the performance of most graph-based clustering methods. To examine the effect of $K$, we design a $K$-sensitivity experiment on the BDGP datasets with $K \in \{1,10\}$. It can be seen from Fig.~\ref{subfig:k} that ACC and NMI remarkably increase as $K$ increases from 1 to 2, and then tends to stabilize in a larger range. It proves that our method is not sensitive to $K$ and can obtain a stable clustering even when there are some spurious links in the graphs.

We then investigate the sensitivity of parameters $\alpha$ and $\beta$, and the ACC and NMI results on Coil20 datasets are shown in Fig.~\ref{subfig:para}. From the figure, it can be seen that our method performs acceptably with most parameter combinations and has relative stability.

\subsection{Training Analysis (Q4)}

Fig.~\ref{subfig:stage2} shows the results of different training stage {\romannumeral2}  starting epochs on Caltech101-20, and Fig.~\ref{subfig:train} shows the curves of loss values, the error of relation graph against the number of epochs. It can be observed from these results that: (1) The loss value shows a downward trend overall and decreases quickly in the first 500 epochs. (2) When the second training phase begins, the error rate drops sharply and then stabilises, indicating the effectiveness of the two stages training and the convergence property of ACTIVE. (3) Two-stage training strategy provides significant performance improvements, but the model is not sensitive to the starting epochs of stage {\romannumeral2}. More results for different datasets and stage {\romannumeral2} start epochs are available in the supplementary material.

\section{Conclusion}
In this paper, we rethink unsupervised contrastive learning and propose a cluster-level contrastive learning framework, namely ACTIVE, to handle the partial multi-view clustering problem. Different from the existing unsupervised contrastive learning methods, we suppose that the representations of similar samples (i.e., belonging to the same cluster) should be similar. To achieve cluster-level contrastive learning, relation graphs are constructed based on existence samples and the known inter-instance relationships are transferred to the missing view to build graphs on the missing data. ACTIVE can lift both contrastive learning and missing data inference from instance-level to cluster-level, effectively mitigating the impact of missing data on clustering. Extensive comparison experiments and ablation studies validate the superiority of the model and the effectiveness of each component. 

\newpage
\bibliographystyle{named}
\bibliography{ijcai22}
\end{document}